\documentclass[11pt,a4paper]{article}
\usepackage[hyperref]{ranlp2025}
\usepackage{times}
\usepackage{latexsym}

\usepackage[table]{xcolor}
\usepackage{graphicx}
\usepackage{multirow}
\usepackage{booktabs}
\usepackage{array}

\usepackage{microtype}
\usepackage{defs}
\usepackage{enumitem}

\aclfinalcopy 

\title{Bias in, Bias out: Annotation Bias in Multilingual Large Language Models}

\author{
 \textbf{Xia Cui\textsuperscript{1},
 \textbf{Ziyi Huang\textsuperscript{2}},
 \textbf{Naeemeh Adel\textsuperscript{1}}
 }
 \\
 \textsuperscript{1}Manchester Metropolitan University, Manchester, UK.
 \\
 \texttt{\{x.cui, n.adel\}@mmu.ac.uk}
\\
 \textsuperscript{2}Hubei University, Wuhan, China.
 \\
 \texttt{ziyihuang@hubu.edu.cn}
\\
}
\date{}

\begin{document}
\maketitle

\begin{abstract}
Annotation bias in NLP datasets remains a major challenge for developing multilingual Large Language Models (LLMs), particularly in culturally diverse settings. Bias from task framing, annotator subjectivity, and cultural mismatches can distort model outputs and exacerbate social harms. 
We propose a comprehensive framework for understanding annotation bias, distinguishing among \textit{instruction bias}, \textit{annotator bias}, and \textit{contextual and cultural bias}. We review detection methods (including inter-annotator agreement, model disagreement, and metadata analysis) and highlight emerging techniques such as multilingual model divergence and cultural inference. We further outline proactive and reactive mitigation strategies, including diverse annotator recruitment, iterative guideline refinement, and post-hoc model adjustments. Our contributions include: (1) a structured typology of annotation bias, (2) a comparative synthesis of detection metrics, (3) an ensemble-based bias mitigation approach adapted for multilingual settings, and (4) an ethical analysis of annotation processes. 
Together, these contributions aim to inform the design of more equitable annotation pipelines for LLMs.

\end{abstract}

\section{Introduction}

Large Language Models (LLMs) such as BERT \citep{Devlin:BERT:2018}, T5 \citep{Raffel:JMLR:2020}, Llama \citep{Touvron:Llama:2023} and GPT-4 \citep{Achiam:GPT4:2023} have transformed Natural Language Processing (NLP), achieving state-of-the-art performance across a wide range of tasks. Their success is largely attributed to pre-training on vast, unannotated corpora that enable them to learn powerful representations. However, aligning these models with human values and adapting them for high-stakes applications requires smaller, curated datasets annotated by humans.

This reliance introduces a critical vulnerability. Annotation bias, which refers to systematic distortions introduced during the labelling process, can severely affect model performance, fairness, and generalisation. It may arise from task framing, annotator subjectivity, or cultural mismatches, and its impact is particularly pronounced in multilingual and culturally heterogeneous contexts \citep{Bender:TACL:2018,Plank:EMNLP:2022}. 

The consequences of annotation bias are not hypothetical. For example, models trained to detect toxicity often misclassify African-American Vernacular English (AAVE) as offensive, due to cultural insensitivity in both annotation guidelines and annotator interpretation. Phrases such as ``That’s my nigga” which carry a supportive meaning in AAVE, are frequently labelled as hateful by annotators unfamiliar with the dialect \citep{Sap:ACL:2019}. This highlights how linguistic and cultural assumptions embedded in the annotation process can lead to unjust model behaviour. 

Such failures reflect a broader pattern. When biased annotations are used for training or fine-tuning, models tend to replicate and sometimes amplify these distortions, resulting in both representational harms and disparities in performance across demographic groups \citep{Dodge:EMNLP:2021, Sheng:EMNLP:2019}. Addressing these issues requires critical scrutiny of annotation workflows, with careful attention to cultural and contextual diversity. 

In this paper, we examine the sources and consequences of annotation bias in multilingual LLMs. We propose a typology of annotation bias, encompassing instruction bias, annotator bias, and contextual or cultural bias. We review established and emerging detection methods, including inter-annotator agreement, model disagreement, and multilingual divergence. We adapt Weak Ensemble Learning (WEL) as a reactive mitigation strategy and assess its effectiveness across multilingual and real-world datasets. Finally, we reflect on the ethical and labour implications of annotation work and suggest directions for building more inclusive and context-aware NLP pipelines.

\section{Background and Motivation}

Early annotation practices in NLP were shaped by linguistic theory and typically involved trained experts using detailed, rule-based guidelines. Datasets such as the Penn Treebank~\citep{Marcus:PennTreebank:1993} and FrameNet~\citep{Baker:FrameNet:1998} exemplified this approach, producing consistent annotations at a small to moderate scale. 

As NLP tasks expanded and model complexity increased, the field shifted toward large-scale annotation through crowdsourcing platforms~\citep{Snow:EMNLP:2008}. This approach enabled the creation of widely used datasets like SNLI~\citep{Bowman:EMNLP:2015} and MultiNLI~\citep{Williams:NAACL-HLT:2018}, but introduced new concerns regarding annotation quality, consistency and subjectivity.

The rise of LLMs has further complicated annotation workflows. Today’s datasets often combine expert review, crowdworker input, and semi-automated methods such as model-in-the-loop annotation~\citep{Pawar:CL:2025}. Many target inherently subjective or ambiguous constructs, including helpfulness, safety, or moral alignment~\citep{Monarch:HCIBook:2021, Uma:JAIR:2022}. These tasks are particularly vulnerable to variation across annotators and contexts.   

At the same time, NLP has increasingly embraced multilingual and multimodal benchmarks. Projects such as XTREME \citep{Hu:ICML:2020}, AmericasNLI~\citep{Ebrahimi:ACL:2022}, TextVQA~\citep{Singh:CVPR:2019}, and HowTo100M~\citep{Miech:ICCV:2019} highlight the challenges of applying traditional annotation schemas across languages, cultures, and modalities. Multimodal tasks introduce further complexity through temporality, affective signals, and cross-modal interpretation.

These trends have exposed a structural vulnerability: annotation bias. This includes not only individual annotator subjectivity, but also culturally conditioned assumptions, linguistic mismatches, and platform-mediated incentives \citep{Blodgett:ACL:2020,Plank:EMNLP:2022}. Annotation decisions made on small but influential datasets can propagate through model fine-tuning and evaluation, leading to downstream harms \citep{Dodge:EMNLP:2021}. 

In response, the field has developed ethical documentation frameworks such as \textit{Data Statements}~\citep{Bender:TACL:2018} and \textit{Datasheets for Datasets}~\citep{Gebru:ACM:2021}. These initiatives promote transparency by capturing the dataset’s linguistic, demographic, and procedural context. They represent an important step toward recognising that high-quality, ethical NLP systems begin with well-understood and well-documented data.

\section{Types of Annotation Bias}
\label{sec:annotation-bias}
Annotation bias in NLP arises when human interpretations, cultural assumptions, or task formulations systematically distort labelled data. These biases can affect model learning, especially during fine-tuning and evaluation. In the context of LLMs, annotation bias often originates from multiple sources and compounds over the pipeline. Addressing it requires distinguishing among different types of bias and understanding how they interact.

We categorise annotation bias into three primary types based on its origin: 
\textbf{Instruction Bias} (Section~\ref{sec:instruction-bias}), \textbf{Annotator Bias} (Section~\ref{sec:annotator-bias}), and \textbf{Contextual and Cultural Bias} (Section~\ref{sec:contextual-bias}). These types are not mutually exclusive. In many cases, a biased annotation reflects an interaction among all three. For example, culturally narrow task guidelines (instruction bias) given to a homogeneous annotator pool (annotator bias) tasked with labelling dialectal language (contextual bias) may produce systematically skewed data. Recognising this interplay is essential for designing effective detection and mitigation strategies~\citep{Bender:TACL:2018}.

\subsection{Instruction Bias}\label{sec:instruction-bias}
\textbf{Instruction bias}~\citep{Parmar:EACL:2023} occurs when the design of an annotation task, including the prompt wording, labelling guidelines, or interface, embeds implicit assumptions that shape how annotators interpret or respond. These assumptions can systematically distort the resulting labels. 

A common example appears in sentiment analysis, where annotators are often asked to classify texts as ``positive", ``negative", or ``neutral". These categories overlook cultural and linguistic nuance, such as expressions of irony, ambivalence, or indirectness~\citep{Mohammad:Emotion:2016}. The framing tends to reflect Western emotional norms that do not generalise across diverse populations~\citep{Huang:ACL:2023}. Similarly, toxicity detection tasks have been shown to mislabel minoritised dialects as offensive, in part due to annotation instructions that lack sociolinguistic sensitivity~\citep{Sap:ACL:2019}.

In the context of LLMs, instruction bias is further complicated by the use of prompts in place of formal annotation guidelines. \textit{Zero-shot}~\citep{Wei:ICLR:2022} and \textit{few-shot} prompting~\citep{Schick:TACL:2022} methods often replace expert-designed protocols. These prompts, though brief, function as implicit task instructions and strongly influence model behaviour. Minor changes in phrasing, such as asking ``Is this inappropriate?" versus ``Is this morally wrong?", can lead to significantly different model outputs, especially for subjective or value-laden tasks~\citep{Zhao:ICML:2021, Schick:TACL:2022, He:EMNLP:2024}.

Moreover, prompts are frequently written by researchers or practitioners who come from specific cultural or disciplinary contexts. Their assumptions shape how tasks are framed and what kinds of answers are considered valid. For example, in mental health detection tasks, prompt templates often reflect Western norms of psychological distress.  This reduces model performance on data from under-represented linguistic or cultural groups~\citep{Parmar:EACL:2023,Cui:IJCNN:2024}. Unlike traditional annotation guidelines, prompts are rarely revised or reviewed through participatory validation processes~\citep{Zamfirescu:CHI:2023,Cui:IJCNN:2024}.

\subsection{Annotator Bias}\label{sec:annotator-bias}
\textbf{Annotator bias} arises from the individual or collective predispositions of those performing the labelling. These may include cognitive heuristics, beliefs, social norms, or demographic characteristics. Even when given identical instructions, annotators interpret data differently depending on their personal context.

Subjective annotation tasks such as toxicity detection, moral judgment, or hate speech classification are particularly susceptible to this type of bias~\citep{Sap:ACL:2019,Liu:ACL:2022,Plank:EMNLP:2022}. Aggregation techniques like majority voting can obscure these differences and suppress minority perspectives, especially when annotator diversity is limited \citep{Aroyo:IEEE:2015,Shardlow:ICLR:2022}.

The rise of crowdwork has intensified these challenges. Annotator pools often differ demographically from both the dataset's source community and its intended application domain~\citep{Eickhoff:ACM:2018,Bender:ACM:2021}. As a result, annotations may misinterpret cultural cues, dialectal language, or context-specific emotional tone. Although such variation is not necessarily the result of carelessness, it can introduce systematic distortion, especially when disagreement is treated as noise rather than signal \citep{Cabitza:AAAI:2023}.

\subsection{Contextual and Cultural Bias}\label{sec:contextual-bias}
\textbf{Contextual and cultural bias} occurs when task design and labelling decisions assume a particular worldview, linguistic norm, or social context. It becomes especially pronounced in multilingual and multimodal tasks, where language, meaning, and affective signals vary widely across cultures.

Annotation labels such as “polite”, “supportive”, or “offensive” often fail to translate cleanly across languages or communities \citep{Ponti:EMNLP:2020:xcopa}. Cultural norms shape how people interpret both language and non-verbal cues, including gestures and tone of voice \citep{Lisa:PSPI:2019, Lukac:Natrue:2023}. Without regionally grounded interpretation frameworks, annotators may mislabel visual or emotional content. 

Additionally, most pretraining data is skewed toward English and Western sources. As of 2025, English accounts for nearly half of all indexed web content \citep{Ani:webdata:2025}. This imbalance in data collection reinforces a corresponding bias in annotation practices.

Recent work has emphasised the importance of culturally grounded taxonomies and community consultation for annotation tasks involving identity, emotion, or morality~\citep{Blodgett:ACL:2020,Hutchinson:ACL:2020,Zhou:EMNLP:2023}. Without such grounding, models trained on annotated data risk reproducing narrow, non-representative worldviews.

\section{Impact on Model Behaviour}

Bias introduced during annotation does not remain confined to the dataset. It propagates into the models trained on that data and leads to measurable downstream harms. This phenomenon, often referred to as “bias in, bias out,” is a central concern in machine learning. When annotation processes reflect cultural, social, or demographic distortions, models tend to reproduce those distortions, and in some cases, amplify them~\citep{Dodge:EMNLP:2021}.

One of the most well-documented consequences is performance disparity across demographic groups. A model may perform well on aggregate metrics while underperforming on texts associated with certain identities, dialects, or cultural contexts. For example, commercial gender classification systems have shown much higher error rates for darker-skinned women. This discrepancy can be traced, in part, to unbalanced training data that lacked diverse and properly annotated examples \citep{Buolamwini:PMLR:2018}. Similarly, recidivism prediction tools have displayed racially skewed false positive rates due to historical biases embedded in the labelled data \citep{Dressel:SA:2018}. 

Beyond accuracy gaps, annotation bias also causes representational harm. These occur when models learn to reproduce social stereotypes or unfair associations. For instance, if training labels disproportionately associate “engineer” with men and “nurse” with women, the model may internalise and repeat these biases in downstream tasks such as text generation or summarisation \citep{Sheng:EMNLP:2019}. In a similar way, toxicity detection models trained on biased annotations may misclassify expressions in African-American Vernacular English (AAVE) as hostile or inappropriate \citep{Sap:ACL:2019}.

These harms can be formalised using established fairness metrics. \textbf{Demographic Parity} requires that the rate of positive predictions be equal across groups. \textbf{Equalised Odds} requires that true and false positive rates remain consistent regardless of group membership. Annotation bias undermines these goals. Returning to the AAVE example, if annotators are more likely to label AAVE expressions as toxic, a classifier trained on such data will exhibit a higher false positive rate for Black speakers. This violates Equalised Odds and leads to unfair penalties against specific communities~\citep{Dixon:ACM:2018}.

These examples demonstrate that annotation bias is not a peripheral issue. It directly contributes to systemic failures in LLMs that affect both technical performance and social impact. For this reason, examining and improving annotation practices is a foundational step toward fairer and more reliable NLP systems.
\section{Case Studies and Empirical Evidence}

To illustrate how annotation bias operates in practice, this section presents two case studies. The first addresses multilingual hate speech detection, where cultural definitions of offence lead to misalignment between training data and deployment contexts. The second focuses on multimodal emotion recognition, where non-verbal cues are interpreted differently across cultural frameworks. These cases highlight that bias often arises not from individual prejudice but from structural mismatches between annotation design and communicative diversity.

\subsection{Case Study: Cross-Cultural Hate Speech Detection}

Hate speech detection is highly sensitive to cultural context. What is considered offensive or harmful in one setting may be acceptable or even humorous in another. This presents a serious challenge for creating models intended to generalise across regions and languages.

\citet{Lee:C3NLP:2023} evaluated monolingual hate speech classifiers across cultural contexts by applying models trained on English-language data from the United States to translated data from languages such as Korean and Arabic. The results showed a drop in F1 scores of up to 42\% and a fourfold increase in false negatives. These failures stemmed not from technical flaws in the models themselves but from annotation biases embedded in the source datasets. Several factors contributed to this performance collapse:
\begin{itemize}[noitemsep,topsep=0pt]
    \item Cultural targets vary. Groups and individuals who are frequent targets of hate speech differ between cultures, meaning training data from one country may miss important examples from another.
\end{itemize}

\begin{itemize}[noitemsep,topsep=0pt]
    \item Sociocultural norms shape expression. Sarcasm, irony, and rhetorical devices have different meanings and social functions depending on the culture.
\end{itemize}

\begin{itemize}[noitemsep,topsep=0pt]
    \item Standards of offensiveness diverge. A statement considered hateful in one community may be seen as neutral or even acceptable in another, depending on social, political, or historical context.
\end{itemize}

This case demonstrates that hate speech is not a culturally neutral construct. Models built on datasets annotated within a single cultural context may fail when applied elsewhere, even if the language is translated. This failure is not only a limitation of model generalisation but also a direct consequence of annotation bias in the original data.

\subsection{Case Study: Multimodal Emotion Recognition}

Bias in multimodal datasets can be more difficult to detect but equally damaging. Emotion recognition tasks that use audio, video, or gesture data rely on the interpretation of non-verbal cues, which are deeply culturally embedded.

\citet{gunes2007bi} conducted a study where they investigated how physical gestures were interpreted across cultural contexts. They found that a single gesture could signal patience in Egypt, positivity in Greece, and confrontation in Italy. When such data are annotated by individuals unfamiliar with the cultural origin of the gesture, systematic mislabelling is likely.

Cultural variation also affects emoji and facial expression interpretation. \citet{gao2020cultural} showed that annotators from Western cultures rely more on mouth shapes to read emoji emotions, while those from Eastern cultures prioritise the eyes. These perceptual differences result in inconsistent annotations and affect model training when emojis are used as supervision signals.

These findings underscore the importance of culturally grounded annotation frameworks in multimodal NLP. Without them, datasets risk encoding a narrow view of human emotion and interaction, reducing the validity and generalisability of trained models.

\section{Detecting Annotation Bias}
Detecting annotation bias is a crucial step toward mitigating its impact on model training and evaluation. A variety of methods have been proposed to identify systematic patterns of bias in annotated datasets, each with different strengths and limitations.
One common approach is to measure inter-annotator agreement (IAA), using metrics such as Cohen’s $\kappa$~\citep{Smeeton:Kappa:1985}.  
For $N$ instances and $M$ annotators, where $y_i^{(j)} \in \mathcal{Y}$ is annotator $j$’s label for instance $i$, the agreement is defined as: 
\begin{equation}
\kappa_{\text{Cohen}} = \frac{p_o - p_e}{1 - p_e}, \quad p_o = \frac{1}{N} \sum_{i=1}^N \mathbb{I}(y_i^{(1)} = y_i^{(2)})
\end{equation}

Here $p_o$ is the observed agreement, i.e., the probability that both annotators assign the \textit{same} label to a randomly selected item. $\mathbb{I}(\cdot)$ is the indicator function. The expected agreement by chance $p_e$ is computed as:
\begin{equation}
p_e = \sum_{k \in \mathcal{Y}} P(y^{(1)} = k) \cdot P(y^{(2)} = k)
\end{equation}
where $P(y^{(j)} = k)$ is the empirical probability of annotator $j$ assigning label $k$.  
Fleiss’ $\kappa$ generalises this metric for multiple annotators~\citep{Fleiss:book:2013}:
\begin{equation}
\kappa_{\text{Fleiss}} = \frac{\bar{p} - \bar{p}_e}{1 - \bar{p}_e}, 
\bar{p} = \frac{1}{N} \sum_{i=1}^N \frac{\sum_{k=1}^K n_{ik}(n_{ik}-1)}{M(M-1)}
\end{equation}
where $n_{ik}$ is the number of annotators assigning label $k$ to instance $i$.  


For settings with missing data or mixed label types, Krippendorff’s $\alpha$~\citep{Krippendorff:phd:2011} offers a more general reliability metric:
\begin{equation}
\alpha = 1 - \frac{D_o}{D_e}
\end{equation}
where $D_o$ is the observed disagreement (weighted across annotator pairs per item) and $D_e$ is the expected disagreement under chance.

A complementary approach is to analyse \textit{model disagreement}. When two models are trained on the same data, divergence in their predictions can reveal annotation ambiguity or bias~\citep{Geva:IJCNLP:2019}.  
For two models $f_1$ and $f_2$, the disagreement rate (DR) is defined as:
\begin{equation}
\text{DR} =
\frac{1}{|\mathcal{D}|} \sum_{x \in \mathcal{D}} \mathbb{I}\big( f_1(x) \neq f_2(x) \big)
\end{equation}
\citet{Uma:JAIR:2022} extend this idea by comparing model predictions with human labels:
\begin{equation}
\Delta =
\frac{1}{|\mathcal{D}|} \sum_{x \in \mathcal{D}} \big| f(x) - y_{\text{human}}(x) \big|
\end{equation}
This metric helps identify inconsistencies between model behaviour and annotation patterns~\citep{Dsouza:COMEDI:2025}.

Another lens on bias detection comes from \textbf{metadata analysis}. By examining annotator demographics, task context, and label distributions, researchers can uncover systematic bias~\citep{Sap:ACL:2019}. For an annotator group $a$, a demographic gap $G(a)$ can be computed as:
\begin{equation}
G(a) =
\left|
\frac{1}{|\mathcal{D}_a|} \sum_{x \in \mathcal{D}_a} y(x) -
\frac{1}{|\mathcal{D}|} \sum_{x \in \mathcal{D}} y(x)
\right|
\end{equation}
Here, $\mathcal{D}_a$ denotes the subset of data annotated by group $a$, and $y(x)$ is the label assigned to instance $x$. A high $G(a)$ may signal systematic differences in annotation patterns between group $a$ and the overall dataset, potentially reflecting underlying biases or cultural variation.

This gap measures how far the group’s average labels deviate from the global average, which may indicate bias or representational disparity~\citep{Sap:ACL:2019}. Traditional metrics, however, may be less effective in multilingual and culturally diverse settings. In these cases, disagreement may reflect true variation rather than annotation error~\citep{Naous:ACL:2024}.  
To address this, new strategies are emerging. \textit{multilingual model disagreement} compares the predictions of models fine-tuned in different languages on parallel corpora $\mathcal{D}_{l_1, l_2}$:
\begin{equation}
\text{DR}(l_1, l_2) =
\frac{1}{|\mathcal{D}_{l_1, l_2}|} 
\sum_{x \in \mathcal{D}_{l_1, l_2}} 
\mathbb{I} \big( f_{l_1}(x) \neq f_{l_2}(x) \big)
\end{equation}
where $f_{l_1}$ and $f_{l_2}$ denote the models fine-tuned in languages $l_1$ and $l_2$, respectively.

Similarly, cultural inference techniques~\citep{Zhang:TGIS:2020,Huang:ACL:2023} use embeddings or sociolinguistic metadata to detect alignment between annotations and cultural backgrounds. 
One such indicator, $\Phi_{\text{cultural}}$ is calculated as the $\ell_2$ distance between two groups:
\begin{equation}
\Phi_{\text{cultural}} =
\left\lVert \, \phi(\mathcal{D}_a) - \phi(\mathcal{D}_{a'}) \, \right\rVert_2 
\end{equation}
where $\phi(\cdot)$ maps a dataset to its cultural embedding space, and $\mathcal{D}_a$, $\mathcal{D}_{a'}$ denote datasets annotated by cultural groups $a$ and $a'$, respectively. 

Together, these methods offer a toolkit for identifying annotation bias at different levels: label consistency, annotator disagreement, cultural framing, and model interpretation. In practice, combining quantitative metrics with qualitative analysis offers the best chance of uncovering and addressing complex forms of annotation bias.



\section{Mitigation Strategies}\label{sec:mitigation}
Detecting annotation bias is only the first step toward creating fair and reliable NLP systems. Effective mitigation requires both proactive strategies, which aim to prevent bias during data collection, and reactive strategies, which address it after annotation or model training. This section outlines techniques across both categories, integrating recent formal approaches with practical best practices.

\subsection{Proactive Strategies}

Proactive strategies aim to reduce annotation bias at the source by redesigning annotation processes with awareness of potential pitfalls.

\paragraph{Diverse Annotator Pools}
To counter annotator bias, it is essential to recruit annotators from a broad range of demographic, cultural, and linguistic backgrounds~\citep{Bender:ACM:2021,Paullada:PAT:2021}. A diverse pool can reveal meaningful disagreements and represent underreported perspectives~\citep{Aroyo:IEEE:2015}. 
One way to quantify diversity is through the entropy of the demographic distribution:
\begin{equation}
H(A) = - \sum_{a \in \mathcal{A}} p(a) \log p(a)
\end{equation}
where $ \mathcal{A} $ is the set of annotator groups and $ p(a) $ is the proportion of annotations from group $a$. A higher entropy score $H(A)$ indicates a more balanced and inclusive annotation pool.


\paragraph{Dynamic Annotation Guidelines}
To mitigate instruction bias, guidelines and prompts should be piloted, reviewed and refined iteratively. This feedback loop helps remove culturally specific assumptions and linguistic ambiguities~\citep{Parmar:EACL:2023}. In LLM-based settings, prompt engineering should be evaluated across cultural contexts to ensure validity~\citep{Zamfirescu:CHI:2023}.  
One can formalise this iterative process by tracking the variance in annotator disagreement across iterations:
\begin{equation}
   \sigma^2_t = \frac{1}{|\mathcal{D}|} \sum_{x \in \mathcal{D}} \text{Var}\big( \{ y^{(t)}_i(x) \}_{i=1}^n \big)
\end{equation}
where $ y^{(t)}_i(x) $ is the label from annotator $i$ on item $x$ during iteration $t$, with the goal that $ \sigma^2_t \to \text{min} $ over $t$.


\paragraph{Culturally Grounded Taxonomies}  
To address contextual and cultural bias, annotation schemes should be developed with culturally grounded taxonomies of emotion, politeness, morality, and related constructs~\citep{Blodgett:ACL:2020,Hutchinson:ACL:2020,Zhou:EMNLP:2023}. Engaging with communities or domain experts helps ensure that annotation labels are valid across languages and cultural settings~\citep{Ponti:EMNLP:2020:xcopa,Naous:ACL:2024}.

\subsection{Reactive Strategies}
Reactive strategies are applied after biases have entered the data or the model. They aim to mitigate downstream harms without necessarily revising the annotation process itself. One key challenge in post-hoc mitigation is handling inconsistencies introduced by annotator subjectivity and instruction bias, particularly when labels reflect divergent interpretations of subjective or culturally loaded concepts. Weighted ensemble methods can address this by leveraging multiple model perspectives to smooth over annotation noise, while still preserving minority viewpoints. 

\paragraph{Post-hoc Model Adjustment}
Biases in trained models can sometimes be mitigated using post-hoc correction methods such as embedding debiasing or output regularisation.~\citet{Kaneko:AACL-IJCNLP:2023} proposed modifying model outputs by subtracting a learned bias component:
\begin{equation}
    f^{\text{debias}}(x) = f_\theta(x) - \lambda b(x)
\end{equation}
where $f_\theta(x)$ is the original model output, $b(x)$ is a bias projection and $\lambda$ controls debiasing strength.


\paragraph{Fine-tuning and In-context Debiasing}
Recent work has explored using targeted fine-tuning~\citep{Webster:arxiv:2021} or in-context prompting~\cite {Ganguli:arxiv:2023} to reshape model behaviour without altering the training data~\citep{Kaneko:COLING:2025}. In the fine-tuning case, model parameters $\theta$ are updated using reweighted or re-annotated dataset $\mathcal{D}^*$:
\begin{equation}
    \min_\theta \mathbb{E}_{(x,y^*) \sim \mathcal{D}^*} \mathcal{L}(f_\theta(x), y^*)
\end{equation}
 
In in-context learning, models are conditioned on carefully constructed prompts $P$ that reduce bias:
\begin{equation}
    f_\theta(y \mid x, P)
\end{equation}
where $P$ is designed to reduce the likelihood of biased completions while preserving task accuracy.


\paragraph{Multi-Objective Weighted Ensemble Learning}
Another reactive strategy leverages ensemble learning to mitigate annotation bias by explicitly modelling annotator disagreement~\citep{Geva:IJCNLP:2019}.  
Given a dataset $\mathcal{D} = \{(x_i, \{y_{i}^{(j)}\}_{j=1}^{M})\}_{i=1}^{N}$, \citet{Ziyi:EMNLP:2025} proposed Weak Ensemble Learning (WEL), which samples one annotator label per instance to construct $K$ label-variant datasets. Each trains a weak predictor $f_{\theta_k}$, weighted by its held-out performance (e.g., F1, cross-entropy, Manhattan distance), with $\sum_{k=1}^K w_k = 1$.  
We extend WEL to a multilingual setting by applying the same label-sampling procedure across datasets in different languages using a shared multilingual model. Final predictions are computed as:
\begin{align}
\hat{y}_i = \sum_{k=1}^K w_k \, f_{\theta_k}(x_i),
\end{align}
allowing the ensemble to capture annotator disagreement while leveraging multilingual representations from a single model.

We use \texttt{mBERT}~\citep{Devlin:BERT:2018} as the base model. 
On the multi-source benchmark from the LeWiDi 2023 shared task~\citep{LeWiDi:SemEval:2023}, WEL generally outperforms baselines using single-model CE loss (CE-only)~\citep{Uma:AAAI:2020} and majority-vote ensembles of top five annotators (Top-5-Ann)~\citep{Xu:ICNLSP:2024}, achieving higher F1 and lower CE/MD scores. The only exception is \textit{ArMIS}, where the very small annotator pool (three annotators) limits the effectiveness of random label sampling.
As the primary focus of this paper is on the discussion of annotation bias in multilingual LLMs, we include the full experimental results in Appendix~\ref{appendix:wel}.






\section{Ethical and Practical Considerations}

The discussion of annotation bias is incomplete without considering the ethical and practical realities of the annotation process itself. Creating high-quality labelled data is not only a technical challenge but also a form of labour that carries human and institutional consequences. These concerns are directly tied to the emergence and persistence of annotation bias because they influence how data are produced, who produces it, and under what conditions.

\subsection{Annotator Wellbeing and Psychological Safety}
One of the most pressing concerns involves the well-being of annotators, particularly those responsible for labelling harmful, toxic, or distressing content. Content moderation datasets, which are essential for training safety filters in LLMs, often expose annotators to a continuous stream of violent, hateful, or traumatic material. Research shows that prolonged exposure to such content can lead to severe psychological effects, including anxiety, depression, insomnia, and symptoms of post-traumatic stress disorder (PTSD)~\citep{das2020fast}. 

This phenomenon is referred to as \textit{vicarious trauma}, a condition in which individuals who are indirectly exposed to trauma begin to show symptoms similar to those of direct trauma survivors~\citep{pearlman1995trauma}.
These effects are compounded by stressful work environments. Annotators often face tight deadlines and high task volumes, with limited autonomy or support systems~\citep{spence2023psychological}. In many cases, stigma around mental health further prevents them from seeking help~\citep{bergman2023overcoming}.

To mitigate these harms, researchers and data curators have a responsibility to implement safeguards. These may include access to mental health services, task rotation to reduce exposure to distressing material, and policies that allow annotators to opt out of specific assignments. Regular breaks, content warnings, and workplace cultures that promote psychological safety are also important steps toward ethical annotation pipelines~\citep{spence2023psychological}.

\subsection{Power Dynamics in Data Labour}
Annotation work is often conducted through crowdworking platforms that rely on a globally distributed, low-cost labour force. These platforms are sometimes described as democratising access to work, but they often reflect significant power asymmetries between requesters and workers.
Annotators frequently operate as anonymous contractors with no job security, limited bargaining power, and little visibility into how their work is used~\citep{Roberts:InBook:2016}. Compensation is usually task-based, which creates incentives to prioritise speed over accuracy. 

This trade-off can result in lower-quality labels and increase the risk of bias in the final dataset~\citep{snow2008cheap}.
Additionally, annotators rarely have channels for providing feedback about unclear instructions, ambiguous data, or annotation policies. As a result, a valuable feedback loop for improving annotation guidelines is often lost~\citep{miceli2022data}.

These structural imbalances are not only ethical concerns; they also have technical implications. Poor working conditions can degrade data quality, obscure disagreement patterns, and exclude minority perspectives~\citep{snow2008cheap}. Creating fairer and more collaborative annotation systems, where annotators are treated as skilled contributors instead of disposable labour, can help ensure both ethical integrity and model reliability.

Ethical considerations must not be separated from methodological concerns. The conditions under which data are created shape their reliability, fairness, and downstream utility. Addressing annotation bias requires attention not only to technical design, but also to the social and economic contexts in which annotation work occurs.
\section{Conclusion and Future Directions}

Annotation bias remains a central challenge for multilingual and multimodal LLMs, shaping how models learn, generalise, and interact with diverse users.  
Mitigation requires both proactive measures (e.g., diverse annotators, refined guidelines) and reactive tools (e.g., bias detection, post-hoc adjustment), underpinned by ethical commitments to annotator well-being and fair labour.  

Future work should prioritise community-driven annotation in marginalised contexts, culturally grounded benchmarks, and richer annotator metadata to improve fairness diagnostics, particularly in low-resource settings. LLMs themselves can assist as scalable annotation and bias-detection tools, but must be guided by real-world social and cultural contexts.  

This paper contributes a typology of annotation bias, surveys detection methods across multilingual and cultural settings, and outlines mitigation strategies. We extend an ensemble-based method to multilingual settings to address label noise and inter-annotator disagreement, demonstrating its effectiveness on four socially sensitive tasks. Incorporating cultural awareness and accountability throughout the data pipeline will help NLP systems better reflect the diversity of human communication.


\section*{Acknowledgments}
We thank the anonymous reviewers and program chairs for their insightful comments and constructive suggestions.

\bibliographystyle{acl_natbib}
\bibliography{datainfo,extra}

\begin{thebibliography}{74}
\expandafter\ifx\csname natexlab\endcsname\relax\def\natexlab#1{#1}\fi

\bibitem[{Achiam et~al.(2023)Achiam, Adler, Agarwal, Ahmad, Akkaya, Aleman, Almeida, Altenschmidt, Altman, Anadkat et~al.}]{Achiam:GPT4:2023}
Josh Achiam, Steven Adler, Sandhini Agarwal, Lama Ahmad, Ilge Akkaya, Florencia~Leoni Aleman, Diogo Almeida, Janko Altenschmidt, Sam Altman, Shyamal Anadkat, et~al. 2023.
\newblock Gpt-4 technical report.
\newblock \emph{arXiv preprint arXiv:2303.08774}.

\bibitem[{{Ani Petrosyan}(2025)}]{Ani:webdata:2025}
{Ani Petrosyan}. 2025.
\newblock \href {https://www.statista.com/statistics/262946/most-common-languages-on-the-internet/} {Most used languages online by share of websites 2025}.

\bibitem[{Aroyo and Welty(2015)}]{Aroyo:IEEE:2015}
Lora Aroyo and Chris Welty. 2015.
\newblock Truth is a lie: Crowd truth and the seven myths of human annotation.
\newblock \emph{AI Mag.}, 36(1):15–24.

\bibitem[{Baker et~al.(1998)Baker, Fillmore, and Lowe}]{Baker:FrameNet:1998}
Collin~F Baker, Charles~J Fillmore, and John~B Lowe. 1998.
\newblock The berkeley framenet project.
\newblock In \emph{COLING 1998 Volume 1: The 17th International Conference on Computational Linguistics}.

\bibitem[{Barrett et~al.(2019)Barrett, Adolphs, Marsella, Martinez, and Pollak}]{Lisa:PSPI:2019}
Lisa~Feldman Barrett, Ralph Adolphs, Stacy Marsella, Aleix~M. Martinez, and Seth~D. Pollak. 2019.
\newblock Emotional expressions reconsidered: Challenges to inferring emotion from human facial movements.
\newblock \emph{Psychological Science in the Public Interest}, 20(1):1--68.
\newblock PMID: 31313636.

\bibitem[{Bender and Friedman(2018)}]{Bender:TACL:2018}
Emily~M. Bender and Batya Friedman. 2018.
\newblock Data statements for natural language processing: Toward mitigating system bias and enabling better science.
\newblock \emph{Transactions of the Association for Computational Linguistics}, 6:587--604.

\bibitem[{Bender et~al.(2021)Bender, Gebru, McMillan-Major, and Shmitchell}]{Bender:ACM:2021}
Emily~M. Bender, Timnit Gebru, Angelina McMillan-Major, and Shmargaret Shmitchell. 2021.
\newblock On the dangers of stochastic parrots: Can language models be too big?
\newblock In \emph{Proceedings of the 2021 ACM Conference on Fairness, Accountability, and Transparency}, FAccT '21, page 610–623, New York, NY, USA. Association for Computing Machinery.

\bibitem[{Bergman and Rushton(2023)}]{bergman2023overcoming}
Alanna Bergman and Cynda~Hylton Rushton. 2023.
\newblock Overcoming stigma: Asking for and receiving mental health support.
\newblock \emph{AACN advanced critical care}, 34(1):67--71.

\bibitem[{Blodgett et~al.(2020)Blodgett, Barocas, Daum{\'e}~III, and Wallach}]{Blodgett:ACL:2020}
Su~Lin Blodgett, Solon Barocas, Hal Daum{\'e}~III, and Hanna Wallach. 2020.
\newblock Language (technology) is power: A critical survey of ``bias'' in {NLP}.
\newblock In \emph{Proceedings of the 58th Annual Meeting of the Association for Computational Linguistics}, pages 5454--5476, Online. Association for Computational Linguistics.

\bibitem[{Bowman et~al.(2015)Bowman, Angeli, Potts, and Manning}]{Bowman:EMNLP:2015}
Samuel~R. Bowman, Gabor Angeli, Christopher Potts, and Christopher~D. Manning. 2015.
\newblock A large annotated corpus for learning natural language inference.
\newblock In \emph{Proceedings of the 2015 Conference on Empirical Methods in Natural Language Processing}, pages 632--642, Lisbon, Portugal. Association for Computational Linguistics.

\bibitem[{Buolamwini and Gebru(2018)}]{Buolamwini:PMLR:2018}
Joy Buolamwini and Timnit Gebru. 2018.
\newblock Gender shades: Intersectional accuracy disparities in commercial gender classification.
\newblock In \emph{Proceedings of the 1st Conference on Fairness, Accountability and Transparency}, volume~81 of \emph{Proceedings of Machine Learning Research}, pages 77--91. PMLR.

\bibitem[{Cabitza et~al.(2023)Cabitza, Campagner, and Basile}]{Cabitza:AAAI:2023}
Federico Cabitza, Andrea Campagner, and Valerio Basile. 2023.
\newblock Toward a perspectivist turn in ground truthing for predictive computing.
\newblock \emph{Proceedings of the AAAI Conference on Artificial Intelligence}, 37(6):6860--6868.

\bibitem[{Cui et~al.(2024)Cui, Hanley, Choudhury, and Mu}]{Cui:IJCNN:2024}
Xia Cui, Terry Hanley, Muj Choudhury, and Tingting Mu. 2024.
\newblock Data-driven or dataless? detecting indicators of mental health difficulties and negative life events in financial resilience using prompt-based learning.
\newblock In \emph{2024 International Joint Conference on Neural Networks (IJCNN)}, pages 1--8.

\bibitem[{Das et~al.(2020)Das, Dang, and Lease}]{das2020fast}
Anubrata Das, Brandon Dang, and Matthew Lease. 2020.
\newblock Fast, accurate, and healthier: Interactive blurring helps moderators reduce exposure to harmful content.
\newblock In \emph{Proceedings of the AAAI conference on human computation and crowdsourcing}, volume~8, pages 33--42.

\bibitem[{Devlin et~al.(2019)Devlin, Chang, Lee, and Toutanova}]{Devlin:BERT:2018}
Jacob Devlin, Ming-Wei Chang, Kenton Lee, and Kristina Toutanova. 2019.
\newblock {BERT}: Pre-training of deep bidirectional transformers for language understanding.
\newblock In \emph{Proceedings of the 2019 Conference of the North {A}merican Chapter of the Association for Computational Linguistics: Human Language Technologies, Volume 1 (Long and Short Papers)}, pages 4171--4186, Minneapolis, Minnesota. Association for Computational Linguistics.

\bibitem[{Dixon et~al.(2018)Dixon, Li, Sorensen, Thain, and Vasserman}]{Dixon:ACM:2018}
Lucas Dixon, John Li, Jeffrey Sorensen, Nithum Thain, and Lucy Vasserman. 2018.
\newblock Measuring and mitigating unintended bias in text classification.
\newblock In \emph{Proceedings of the 2018 AAAI/ACM Conference on AI, Ethics, and Society}, AIES '18, page 67–73, New York, NY, USA. Association for Computing Machinery.

\bibitem[{Dodge et~al.(2021)Dodge, Sap, Marasovi{\'c}, Agnew, Ilharco, Groeneveld, Mitchell, and Gardner}]{Dodge:EMNLP:2021}
Jesse Dodge, Maarten Sap, Ana Marasovi{\'c}, William Agnew, Gabriel Ilharco, Dirk Groeneveld, Margaret Mitchell, and Matt Gardner. 2021.
\newblock Documenting large webtext corpora: A case study on the colossal clean crawled corpus.
\newblock In \emph{Proceedings of the 2021 Conference on Empirical Methods in Natural Language Processing}, pages 1286--1305, Online and Punta Cana, Dominican Republic. Association for Computational Linguistics.

\bibitem[{Dressel and Farid(2018)}]{Dressel:SA:2018}
Julia Dressel and Hany Farid. 2018.
\newblock The accuracy, fairness, and limits of predicting recidivism.
\newblock \emph{Science Advances}, 4(1):eaao5580.

\bibitem[{Dsouza and Kovatchev(2025)}]{Dsouza:COMEDI:2025}
Russel Dsouza and Venelin Kovatchev. 2025.
\newblock Sources of disagreement in data for {LLM} instruction tuning.
\newblock In \emph{Proceedings of Context and Meaning: Navigating Disagreements in NLP Annotation}, pages 20--32, Abu Dhabi, UAE. International Committee on Computational Linguistics.

\bibitem[{Ebrahimi et~al.(2022)Ebrahimi, Mager, Oncevay, Chaudhary, Chiruzzo, Fan, Ortega, Ramos, Rios, Meza~Ruiz, Gim{\'e}nez-Lugo, Mager, Neubig, Palmer, Coto-Solano, Vu, and Kann}]{Ebrahimi:ACL:2022}
Abteen Ebrahimi, Manuel Mager, Arturo Oncevay, Vishrav Chaudhary, Luis Chiruzzo, Angela Fan, John Ortega, Ricardo Ramos, Annette Rios, Ivan~Vladimir Meza~Ruiz, Gustavo Gim{\'e}nez-Lugo, Elisabeth Mager, Graham Neubig, Alexis Palmer, Rolando Coto-Solano, Thang Vu, and Katharina Kann. 2022.
\newblock {A}mericas{NLI}: Evaluating zero-shot natural language understanding of pretrained multilingual models in truly low-resource languages.
\newblock In \emph{Proceedings of the 60th Annual Meeting of the Association for Computational Linguistics (Volume 1: Long Papers)}, pages 6279--6299, Dublin, Ireland. Association for Computational Linguistics.

\bibitem[{Eickhoff(2018)}]{Eickhoff:ACM:2018}
Carsten Eickhoff. 2018.
\newblock Cognitive biases in crowdsourcing.
\newblock In \emph{Proceedings of the Eleventh ACM International Conference on Web Search and Data Mining}, WSDM '18, page 162–170, New York, NY, USA. Association for Computing Machinery.

\bibitem[{Fleiss et~al.(2013)Fleiss, Levin, and Paik}]{Fleiss:book:2013}
Joseph~L Fleiss, Bruce Levin, and Myunghee~Cho Paik. 2013.
\newblock \emph{Statistical methods for rates and proportions}.
\newblock john wiley \& sons.

\bibitem[{Ganguli et~al.(2023)Ganguli, Askell, Schiefer, Liao, Lukošiūtė, Chen, Goldie, Mirhoseini, Olsson, Hernandez, Drain, Li, Tran-Johnson, Perez, Kernion, Kerr, Mueller, Landau, Ndousse, Nguyen, Lovitt, Sellitto, Elhage, Mercado, DasSarma, Rausch, Lasenby, Larson, Ringer, Kundu, Kadavath, Johnston, Kravec, Showk, Lanham, Telleen-Lawton, Henighan, Hume, Bai, Hatfield-Dodds, Mann, Amodei, Joseph, McCandlish, Brown, Olah, Clark, Bowman, and Kaplan}]{Ganguli:arxiv:2023}
Deep Ganguli, Amanda Askell, Nicholas Schiefer, Thomas~I. Liao, Kamilė Lukošiūtė, Anna Chen, Anna Goldie, Azalia Mirhoseini, Catherine Olsson, Danny Hernandez, Dawn Drain, Dustin Li, Eli Tran-Johnson, Ethan Perez, Jackson Kernion, Jamie Kerr, Jared Mueller, Joshua Landau, Kamal Ndousse, Karina Nguyen, Liane Lovitt, Michael Sellitto, Nelson Elhage, Noemi Mercado, Nova DasSarma, Oliver Rausch, Robert Lasenby, Robin Larson, Sam Ringer, Sandipan Kundu, Saurav Kadavath, Scott Johnston, Shauna Kravec, Sheer~El Showk, Tamera Lanham, Timothy Telleen-Lawton, Tom Henighan, Tristan Hume, Yuntao Bai, Zac Hatfield-Dodds, Ben Mann, Dario Amodei, Nicholas Joseph, Sam McCandlish, Tom Brown, Christopher Olah, Jack Clark, Samuel~R. Bowman, and Jared Kaplan. 2023.
\newblock \href {http://arxiv.org/abs/2302.07459} {The capacity for moral self-correction in large language models}.

\bibitem[{Gao and VanderLaan(2020)}]{gao2020cultural}
Boting Gao and Doug~P VanderLaan. 2020.
\newblock Cultural influences on perceptions of emotions depicted in emojis.
\newblock \emph{Cyberpsychology, Behavior, and Social Networking}, 23(8):567--570.

\bibitem[{Gebru et~al.(2021)Gebru, Morgenstern, Vecchione, Vaughan, Wallach, III, and Crawford}]{Gebru:ACM:2021}
Timnit Gebru, Jamie Morgenstern, Briana Vecchione, Jennifer~Wortman Vaughan, Hanna Wallach, Hal~Daum\'{e} III, and Kate Crawford. 2021.
\newblock Datasheets for datasets.
\newblock \emph{Commun. ACM}, 64(12):86–92.

\bibitem[{Geva et~al.(2019)Geva, Goldberg, and Berant}]{Geva:IJCNLP:2019}
Mor Geva, Yoav Goldberg, and Jonathan Berant. 2019.
\newblock Are we modeling the task or the annotator? an investigation of annotator bias in natural language understanding datasets.
\newblock In \emph{Proceedings of the 2019 Conference on Empirical Methods in Natural Language Processing and the 9th International Joint Conference on Natural Language Processing (EMNLP-IJCNLP)}, pages 1161--1166, Hong Kong, China. Association for Computational Linguistics.

\bibitem[{Gunes and Piccardi(2007)}]{gunes2007bi}
Hatice Gunes and Massimo Piccardi. 2007.
\newblock Bi-modal emotion recognition from expressive face and body gestures.
\newblock \emph{Journal of Network and Computer Applications}, 30(4):1334--1345.

\bibitem[{He et~al.(2024)He, Long, and Roy}]{He:EMNLP:2024}
Kang He, Yinghan Long, and Kaushik Roy. 2024.
\newblock Prompt-based bias calibration for better zero/few-shot learning of language models.
\newblock In \emph{Findings of the Association for Computational Linguistics: EMNLP 2024}, pages 12673--12691, Miami, Florida, USA. Association for Computational Linguistics.

\bibitem[{Hu et~al.(2020)Hu, Ruder, Siddhant, Neubig, Firat, and Johnson}]{Hu:ICML:2020}
Junjie Hu, Sebastian Ruder, Aditya Siddhant, Graham Neubig, Orhan Firat, and Melvin Johnson. 2020.
\newblock Xtreme: a massively multilingual multi-task benchmark for evaluating cross-lingual generalization.
\newblock In \emph{Proceedings of the 37th International Conference on Machine Learning}, ICML'20. JMLR.org.

\bibitem[{Huang and Yang(2023)}]{Huang:ACL:2023}
Jing Huang and Diyi Yang. 2023.
\newblock Culturally aware natural language inference.
\newblock In \emph{Findings of the Association for Computational Linguistics: EMNLP 2023}, pages 7591--7609, Singapore. Association for Computational Linguistics.

\bibitem[{Huang et~al.(2025)Huang, Abeynayake, and Cui}]{Ziyi:EMNLP:2025}
Ziyi Huang, Nishanthi~Rupika Abeynayake, and Xia Cui. 2025.
\newblock Weak ensemble learning from multiple annotators for subjective text classification.
\newblock In \emph{Proceedings of the 4th Workshop on Perspectivist Approaches to NLP (NLPerspectives) @ EMNLP 2025}, Suzhou, China. Association for Computational Linguistics.

\bibitem[{Hutchinson et~al.(2020)Hutchinson, Prabhakaran, Denton, Webster, Zhong, and Denuyl}]{Hutchinson:ACL:2020}
Ben Hutchinson, Vinodkumar Prabhakaran, Emily Denton, Kellie Webster, Yu~Zhong, and Stephen Denuyl. 2020.
\newblock Social biases in {NLP} models as barriers for persons with disabilities.
\newblock In \emph{Proceedings of the 58th Annual Meeting of the Association for Computational Linguistics}, pages 5491--5501, Online. Association for Computational Linguistics.

\bibitem[{Kaneko et~al.(2025)Kaneko, Bollegala, and Baldwin}]{Kaneko:COLING:2025}
Masahiro Kaneko, Danushka Bollegala, and Timothy Baldwin. 2025.
\newblock The gaps between fine tuning and in-context learning in bias evaluation and debiasing.
\newblock In \emph{Proceedings of the 31st International Conference on Computational Linguistics}, pages 2758--2764, Abu Dhabi, UAE. Association for Computational Linguistics.

\bibitem[{Kaneko et~al.(2023)Kaneko, Bollegala, and Okazaki}]{Kaneko:AACL-IJCNLP:2023}
Masahiro Kaneko, Danushka Bollegala, and Naoaki Okazaki. 2023.
\newblock The impact of debiasing on the performance of language models in downstream tasks is underestimated.
\newblock In \emph{Proceedings of the 13th International Joint Conference on Natural Language Processing and the 3rd Conference of the Asia-Pacific Chapter of the Association for Computational Linguistics (Volume 2: Short Papers)}, pages 29--36, Nusa Dua, Bali. Association for Computational Linguistics.

\bibitem[{Krippendorff(2011)}]{Krippendorff:phd:2011}
Klaus Krippendorff. 2011.
\newblock Computing krippendorff's alpha-reliability.

\bibitem[{Lee et~al.(2023)Lee, Jung, and Oh}]{Lee:C3NLP:2023}
Nayeon Lee, Chani Jung, and Alice Oh. 2023.
\newblock Hate speech classifiers are culturally insensitive.
\newblock In \emph{Proceedings of the First Workshop on Cross-Cultural Considerations in NLP (C3NLP)}, pages 35--46, Dubrovnik, Croatia. Association for Computational Linguistics.

\bibitem[{Leonardellli et~al.(2023)Leonardellli, Abercrombie, Almanea, Basile, Fornaciari, Plank, Poesio, Rieser, and Uma}]{LeWiDi:SemEval:2023}
Elisa Leonardellli, Gavin Abercrombie, Dina Almanea, Valerio Basile, Tommaso Fornaciari, Barbara Plank, Massimo Poesio, Verena Rieser, and Alexandra Uma. 2023.
\newblock {SemEval-2023 Task 11: Learning With Disagreements (LeWiDi)}.
\newblock In \emph{Proceedings of the 17th International Workshop on Semantic Evaluation}, Toronto, Canada. {Association for Computational Linguistics}.

\bibitem[{Liu et~al.(2022)Liu, Thekinen, Mollaoglu, Tang, Yang, Cheng, Liu, and Tang}]{Liu:ACL:2022}
Haochen Liu, Joseph Thekinen, Sinem Mollaoglu, Da~Tang, Ji~Yang, Youlong Cheng, Hui Liu, and Jiliang Tang. 2022.
\newblock Toward annotator group bias in crowdsourcing.
\newblock In \emph{Proceedings of the 60th Annual Meeting of the Association for Computational Linguistics (Volume 1: Long Papers)}, pages 1797--1806, Dublin, Ireland. Association for Computational Linguistics.

\bibitem[{Lukac et~al.(2023)Lukac, Zhambulova, Abdiyeva, and Lewis}]{Lukac:Natrue:2023}
Martin Lukac, Gulnaz Zhambulova, Kamila Abdiyeva, and Michael Lewis. 2023.
\newblock Study on emotion recognition bias in different regional groups.
\newblock \emph{Scientific Reports}, 13(1):8414.

\bibitem[{Marcus et~al.(1993)Marcus, Santorini, and Marcinkiewicz}]{Marcus:PennTreebank:1993}
Mitch Marcus, Beatrice Santorini, and Mary~Ann Marcinkiewicz. 1993.
\newblock Building a large annotated corpus of english: The penn treebank.
\newblock \emph{Computational linguistics}, 19(2):313--330.

\bibitem[{Miceli and Posada(2022)}]{miceli2022data}
Milagros Miceli and Julian Posada. 2022.
\newblock The data-production dispositif.
\newblock \emph{Proceedings of the ACM on human-computer interaction}, 6(CSCW2):1--37.

\bibitem[{Miech et~al.(2019)Miech, Zhukov, Alayrac, Tapaswi, Laptev, and Sivic}]{Miech:ICCV:2019}
Antoine Miech, Dimitri Zhukov, Jean-Baptiste Alayrac, Makarand Tapaswi, Ivan Laptev, and Josef Sivic. 2019.
\newblock Howto100m: Learning a text-video embedding by watching hundred million narrated video clips.
\newblock In \emph{Proceedings of the IEEE/CVF International Conference on Computer Vision (ICCV)}.

\bibitem[{Mohammad(2016)}]{Mohammad:Emotion:2016}
Saif~M. Mohammad. 2016.
\newblock 9 - sentiment analysis: Detecting valence, emotions, and other affectual states from text.
\newblock In Herbert~L. Meiselman, editor, \emph{Emotion Measurement}, pages 201--237. Woodhead Publishing.

\bibitem[{Monarch(2021)}]{Monarch:HCIBook:2021}
Robert~Munro Monarch. 2021.
\newblock \emph{Human-in-the-Loop Machine Learning: Active learning and annotation for human-centered AI}.
\newblock Simon and Schuster.

\bibitem[{Naous et~al.(2024)Naous, Ryan, Ritter, and Xu}]{Naous:ACL:2024}
Tarek Naous, Michael~J Ryan, Alan Ritter, and Wei Xu. 2024.
\newblock Having beer after prayer? measuring cultural bias in large language models.
\newblock In \emph{Proceedings of the 62nd Annual Meeting of the Association for Computational Linguistics (Volume 1: Long Papers)}, pages 16366--16393, Bangkok, Thailand. Association for Computational Linguistics.

\bibitem[{Parmar et~al.(2023)Parmar, Mishra, Geva, and Baral}]{Parmar:EACL:2023}
Mihir Parmar, Swaroop Mishra, Mor Geva, and Chitta Baral. 2023.
\newblock Don{'}t blame the annotator: Bias already starts in the annotation instructions.
\newblock In \emph{Proceedings of the 17th Conference of the European Chapter of the Association for Computational Linguistics}, pages 1779--1789, Dubrovnik, Croatia. Association for Computational Linguistics.

\bibitem[{Paullada et~al.(2021)Paullada, Raji, Bender, Denton, and Hanna}]{Paullada:PAT:2021}
Amandalynne Paullada, Inioluwa~Deborah Raji, Emily~M. Bender, Emily Denton, and Alex Hanna. 2021.
\newblock Data and its (dis)contents: A survey of dataset development and use in machine learning research.
\newblock \emph{Patterns}, 2(11):100336.

\bibitem[{Pawar et~al.(2025)Pawar, Park, Jin, Arora, Myung, Yadav, Haznitrama, Song, Oh, and Augenstein}]{Pawar:CL:2025}
Siddhesh Pawar, Junyeong Park, Jiho Jin, Arnav Arora, Junho Myung, Srishti Yadav, Faiz~Ghifari Haznitrama, Inhwa Song, Alice Oh, and Isabelle Augenstein. 2025.
\newblock Survey of cultural awareness in language models: Text and beyond.
\newblock \emph{Computational Linguistics}, pages 1--96.

\bibitem[{Pearlman and Saakvitne(1995)}]{pearlman1995trauma}
Laurie~Anne Pearlman and Karen~W Saakvitne. 1995.
\newblock \emph{Trauma and the therapist: Countertransference and vicarious traumatization in psychotherapy with incest survivors.}
\newblock WW Norton \& Company.

\bibitem[{Plank(2022)}]{Plank:EMNLP:2022}
Barbara Plank. 2022.
\newblock The ``problem'' of human label variation: On ground truth in data, modeling and evaluation.
\newblock In \emph{Proceedings of the 2022 Conference on Empirical Methods in Natural Language Processing}, Abu Dhabi. Association for Computational Linguistics.

\bibitem[{Ponti et~al.(2020)Ponti, Glava{\v{s}}, Majewska, Liu, Vuli{\'c}, and Korhonen}]{Ponti:EMNLP:2020:xcopa}
Edoardo~Maria Ponti, Goran Glava{\v{s}}, Olga Majewska, Qianchu Liu, Ivan Vuli{\'c}, and Anna Korhonen. 2020.
\newblock {XCOPA}: A multilingual dataset for causal commonsense reasoning.
\newblock In \emph{Proceedings of the 2020 Conference on Empirical Methods in Natural Language Processing (EMNLP)}, pages 2362--2376, Online. Association for Computational Linguistics.

\bibitem[{Raffel et~al.(2020)Raffel, Shazeer, Roberts, Lee, Narang, Matena, Zhou, Li, and Liu}]{Raffel:JMLR:2020}
Colin Raffel, Noam Shazeer, Adam Roberts, Katherine Lee, Sharan Narang, Michael Matena, Yanqi Zhou, Wei Li, and Peter~J. Liu. 2020.
\newblock Exploring the limits of transfer learning with a unified text-to-text transformer.
\newblock \emph{Journal of Machine Learning Research}, 21(140):1--67.

\bibitem[{Roberts(2016)}]{Roberts:InBook:2016}
Sarah~T. Roberts. 2016.
\newblock \emph{The Intersectional Internet: Race, Sex, Class and Culture Online}, chapter Commercial content moderation: Digital labourers'dirty work. Peter Lang Publishing.

\bibitem[{Sap et~al.(2019)Sap, Card, Gabriel, Choi, and Smith}]{Sap:ACL:2019}
Maarten Sap, Dallas Card, Saadia Gabriel, Yejin Choi, and Noah~A. Smith. 2019.
\newblock The risk of racial bias in hate speech detection.
\newblock In \emph{Proceedings of the 57th Annual Meeting of the Association for Computational Linguistics}, pages 1668--1678, Florence, Italy. Association for Computational Linguistics.

\bibitem[{Schick and Sch{\"u}tze(2022)}]{Schick:TACL:2022}
Timo Schick and Hinrich Sch{\"u}tze. 2022.
\newblock True few-shot learning with {P}rompts{---}{A} real-world perspective.
\newblock \emph{Transactions of the Association for Computational Linguistics}, 10:716--731.

\bibitem[{Shardlow(2022)}]{Shardlow:ICLR:2022}
Matthew Shardlow. 2022.
\newblock Agree to disagree: Exploring subjectivity in lexical complexity.
\newblock In \emph{Proceedings of the 2nd Workshop on Tools and Resources to Empower People with REAding DIfficulties (READI) within the 13th Language Resources and Evaluation Conference}, pages 9--16, Marseille, France. European Language Resources Association.

\bibitem[{Sheng et~al.(2019)Sheng, Chang, Natarajan, and Peng}]{Sheng:EMNLP:2019}
Emily Sheng, Kai-Wei Chang, Premkumar Natarajan, and Nanyun Peng. 2019.
\newblock The woman worked as a babysitter: On biases in language generation.
\newblock In \emph{Proceedings of the 2019 Conference on Empirical Methods in Natural Language Processing and the 9th International Joint Conference on Natural Language Processing (EMNLP-IJCNLP)}, pages 3407--3412, Hong Kong, China. Association for Computational Linguistics.

\bibitem[{Singh et~al.(2019)Singh, Natarajan, Shah, Jiang, Chen, Batra, Parikh, and Rohrbach}]{Singh:CVPR:2019}
Amanpreet Singh, Vivek Natarajan, Meet Shah, Yu~Jiang, Xinlei Chen, Dhruv Batra, Devi Parikh, and Marcus Rohrbach. 2019.
\newblock Towards vqa models that can read.
\newblock In \emph{Proceedings of the IEEE/CVF Conference on Computer Vision and Pattern Recognition (CVPR)}.

\bibitem[{Smeeton(1985)}]{Smeeton:Kappa:1985}
Nigel~C Smeeton. 1985.
\newblock Early history of the kappa statistic.
\newblock \emph{Biometrics}, 41:795.

\bibitem[{Snow et~al.(2008{\natexlab{a}})Snow, O{'}Connor, Jurafsky, and Ng}]{Snow:EMNLP:2008}
Rion Snow, Brendan O{'}Connor, Daniel Jurafsky, and Andrew Ng. 2008{\natexlab{a}}.
\newblock Cheap and fast {--} but is it good? evaluating non-expert annotations for natural language tasks.
\newblock In \emph{Proceedings of the 2008 Conference on Empirical Methods in Natural Language Processing}, pages 254--263, Honolulu, Hawaii. Association for Computational Linguistics.

\bibitem[{Snow et~al.(2008{\natexlab{b}})Snow, O’connor, Jurafsky, and Ng}]{snow2008cheap}
Rion Snow, Brendan O’connor, Dan Jurafsky, and Andrew~Y Ng. 2008{\natexlab{b}}.
\newblock Cheap and fast--but is it good? evaluating non-expert annotations for natural language tasks.
\newblock In \emph{Proceedings of the 2008 conference on empirical methods in natural language processing}, pages 254--263.

\bibitem[{Spence et~al.(2023)Spence, Bifulco, Bradbury, Martellozzo, and DeMarco}]{spence2023psychological}
Ruth Spence, Antonia Bifulco, Paula Bradbury, Elena Martellozzo, and Jeffrey DeMarco. 2023.
\newblock The psychological impacts of content moderation on content moderators: A qualitative study.
\newblock \emph{Cyberpsychology: Journal of Psychosocial Research on Cyberspace}, 17(4).

\bibitem[{Touvron et~al.(2023)Touvron, Lavril, Izacard, Martinet, Lachaux, Lacroix, Rozi{\`e}re, Goyal, Hambro, Azhar et~al.}]{Touvron:Llama:2023}
Hugo Touvron, Thibaut Lavril, Gautier Izacard, Xavier Martinet, Marie-Anne Lachaux, Timoth{\'e}e Lacroix, Baptiste Rozi{\`e}re, Naman Goyal, Eric Hambro, Faisal Azhar, et~al. 2023.
\newblock Llama: Open and efficient foundation language models.
\newblock \emph{arXiv preprint arXiv:2302.13971}.

\bibitem[{Uma et~al.(2020)Uma, Fornaciari, Hovy, Paun, Plank, and Poesio}]{Uma:AAAI:2020}
Alexandra Uma, Tommaso Fornaciari, Dirk Hovy, Silviu Paun, Barbara Plank, and Massimo Poesio. 2020.
\newblock A case for soft loss functions.
\newblock \emph{Proceedings of the AAAI Conference on Human Computation and Crowdsourcing}, 8(1):173--177.

\bibitem[{Uma et~al.(2022)Uma, Fornaciari, Hovy, Paun, Plank, and Poesio}]{Uma:JAIR:2022}
Alexandra~N. Uma, Tommaso Fornaciari, Dirk Hovy, Silviu Paun, Barbara Plank, and Massimo Poesio. 2022.
\newblock Learning from disagreement: A survey.
\newblock volume~72, page 1385–1470, El Segundo, CA, USA. AI Access Foundation.

\bibitem[{Webster et~al.(2021)Webster, Wang, Tenney, Beutel, Pitler, Pavlick, Chen, Chi, and Petrov}]{Webster:arxiv:2021}
Kellie Webster, Xuezhi Wang, Ian Tenney, Alex Beutel, Emily Pitler, Ellie Pavlick, Jilin Chen, Ed~Chi, and Slav Petrov. 2021.
\newblock \href {http://arxiv.org/abs/2010.06032} {Measuring and reducing gendered correlations in pre-trained models}.

\bibitem[{Wei et~al.(2022)Wei, Bosma, Zhao, Guu, Yu, Lester, Du, Dai, and Le}]{Wei:ICLR:2022}
Jason Wei, Maarten Bosma, Vincent~Y. Zhao, Kelvin Guu, Adams~Wei Yu, Brian Lester, Nan Du, Andrew~M. Dai, and Quoc~V. Le. 2022.
\newblock Finetuned language models are zero-shot learners.
\newblock In \emph{The Tenth International Conference on Learning Representations, ICLR 2022, Virtual Event, April 25-29, 2022}. OpenReview.net.

\bibitem[{Williams et~al.(2018)Williams, Nangia, and Bowman}]{Williams:NAACL-HLT:2018}
Adina Williams, Nikita Nangia, and Samuel Bowman. 2018.
\newblock A broad-coverage challenge corpus for sentence understanding through inference.
\newblock In \emph{Proceedings of the 2018 Conference of the North {A}merican Chapter of the Association for Computational Linguistics: Human Language Technologies, Volume 1 (Long Papers)}, pages 1112--1122, New Orleans, Louisiana. Association for Computational Linguistics.

\bibitem[{Xu et~al.(2024)Xu, Theune, and Braun}]{Xu:ICNLSP:2024}
Jin Xu, Mari{\"e}t Theune, and Daniel Braun. 2024.
\newblock Leveraging annotator disagreement for text classification.
\newblock In \emph{Proceedings of the 7th International Conference on Natural Language and Speech Processing (ICNLSP 2024)}, pages 1--10, Trento. Association for Computational Linguistics.

\bibitem[{Zamfirescu-Pereira et~al.(2023)Zamfirescu-Pereira, Wong, Hartmann, and Yang}]{Zamfirescu:CHI:2023}
J~Diego Zamfirescu-Pereira, Richmond~Y Wong, Bjoern Hartmann, and Qian Yang. 2023.
\newblock Why johnny can’t prompt: how non-ai experts try (and fail) to design llm prompts.
\newblock In \emph{Proceedings of the 2023 CHI conference on human factors in computing systems}, pages 1--21.

\bibitem[{Zhang et~al.(2020{\natexlab{a}})Zhang, Bai, Zhang, Bai, Zhu, and Zhao}]{Zhang:ACL:2020}
Guanhua Zhang, Bing Bai, Junqi Zhang, Kun Bai, Conghui Zhu, and Tiejun Zhao. 2020{\natexlab{a}}.
\newblock Demographics should not be the reason of toxicity: Mitigating discrimination in text classifications with instance weighting.
\newblock In \emph{Proceedings of the 58th Annual Meeting of the Association for Computational Linguistics}, pages 4134--4145, Online. Association for Computational Linguistics.

\bibitem[{Zhang et~al.(2020{\natexlab{b}})Zhang, Zhou, Tang, Ji, Zhang, and Xiong}]{Zhang:TGIS:2020}
Haiping Zhang, Xingxing Zhou, Guoan Tang, Genlin Ji, Xueying Zhang, and Liyang Xiong. 2020{\natexlab{b}}.
\newblock Inference method for cultural diffusion patterns using a field model.
\newblock \emph{Transactions in GIS}, 24(6):1578--1601.

\bibitem[{Zhao et~al.(2021)Zhao, Wallace, Feng, Klein, and Singh}]{Zhao:ICML:2021}
Tony Zhao, Eric Wallace, Shi Feng, Dan Klein, and Sameer Singh. 2021.
\newblock Calibrate before use: Improving few-shot performance of language models.
\newblock In \emph{International Conference on Machine Learning}.

\bibitem[{Zhou et~al.(2023)Zhou, Camacho-Collados, and Bollegala}]{Zhou:EMNLP:2023}
Yi~Zhou, Jose Camacho-Collados, and Danushka Bollegala. 2023.
\newblock A predictive factor analysis of social biases and task-performance in pretrained masked language models.
\newblock In \emph{Proceedings of the 2023 Conference on Empirical Methods in Natural Language Processing}, pages 11082--11100, Singapore. Association for Computational Linguistics.

\end{thebibliography}

\clearpage
\appendix
\section*{Supplementary Materials}
\definecolor{instructionblue}{RGB}{13,71,161}
\definecolor{annotatorgreen}{RGB}{27,94,32}
\definecolor{contextualred}{RGB}{183,28,28}
\definecolor{connectionpurple}{RGB}{123,31,162}
\section{Relationships Among Bias Types, Detection, and Mitigation}\label{appendix:bias_relationships}

\begin{figure*}[tbh]
    \centering
    \includegraphics[width=1\textwidth]{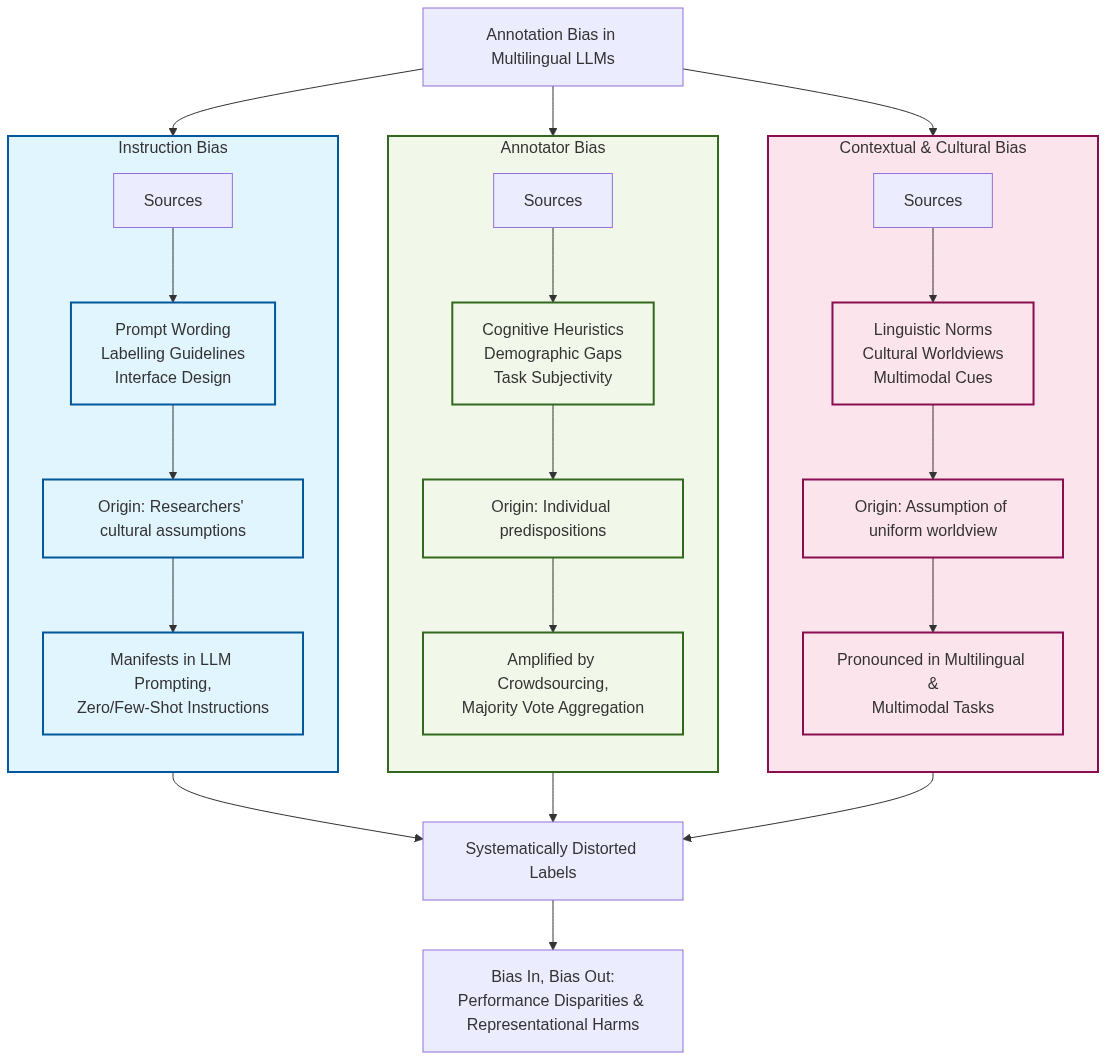}
    \caption{Taxonomy of annotation bias types observed in multilingual LLMs, showing three primary categories with distinct colouring: \textcolor[RGB]{1,87,155}{Instruction Bias}, \textcolor[RGB]{51,105,30}{Annotator Bias}, and \textcolor[RGB]{136,14,79}{Contextual \& Cultural Bias}.}
    \label{fig:bias_taxonomy}
\end{figure*}

This supplementary section details the observed relationships between annotation bias types and their associated detection and mitigation approaches, as illustrated in Figures~\ref{fig:bias_taxonomy} and \ref{fig:bias_detection_relations}. These relationships emerged from our analysis of current literature and empirical findings, rather than constituting a prescriptive framework.

\subsection{Relationships Between Bias Types}

Our analysis identifies three primary bias types that frequently interact in annotation processes:
(1) \textcolor{instructionblue}{\textbf{Instruction Bias}}: Arising from task design choices, guidelines, and prompt formulations;
(2) \textcolor{annotatorgreen}{\textbf{Annotator Bias}}: Stemming from individual predispositions and demographic characteristics;
(3) \textcolor{contextualred}{\textbf{Contextual \& Cultural Bias}}: Emerging from cultural mismatches and linguistic norms.
These bias types often co-occur and compound each other, particularly in multilingual settings where cultural context influences both task interpretation and annotation execution.

\subsection{Correlations Between Detection and Mitigation Approaches}
\begin{figure*}[tbh]
    \centering
    \includegraphics[width=0.9\textwidth]{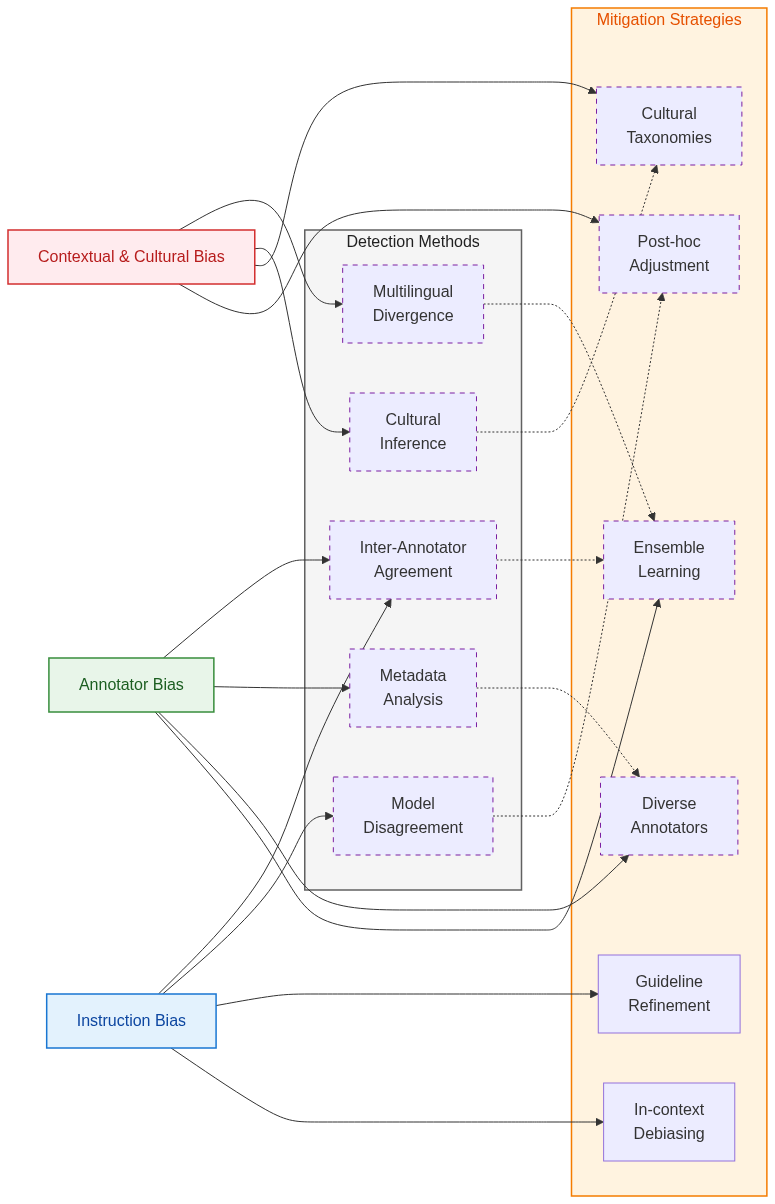}
    \caption{Relationships between annotation bias sources, detection methods, and mitigation strategies. Solid lines indicate primary correlations; dashed lines (\textcolor[RGB]{123,31,162}{purple}) show secondary cross-connections.}
    \label{fig:bias_detection_relations}
\end{figure*}

Figure~\ref{fig:bias_detection_relations} illustrates correlations observed between specific bias types and effective handling strategies.

\textcolor{instructionblue}{Instruction bias} correlations include detection through inter-annotator agreement metrics \citep{Krippendorff:phd:2011} and model disagreement analysis \citep{Geva:IJCNLP:2019}, with mitigation through guideline refinement \citep{Parmar:EACL:2023} and in-context debiasing \citep{Ganguli:arxiv:2023}.
    
\textcolor{annotatorgreen}{Annotator bias} correlations involve detection via metadata analysis \citep{Sap:ACL:2019} with mitigation through diverse annotator recruitment \citep{Bender:TACL:2018} and weak ensemble learning \citep{Ziyi:EMNLP:2025}.
    
\textcolor{contextualred}{Contextual and cultural bias} correlations include detection via multilingual divergence analysis \citep{Huang:ACL:2023} and cultural inference methods \citep{Zhang:ACL:2020}, with mitigation through culturally grounded taxonomies \citep{Ponti:EMNLP:2020:xcopa} and post-hoc adjustments \citep{Kaneko:AACL-IJCNLP:2023}.

\subsection{Emergent Cross-Connections}

Our analysis reveals several emergent cross-connections where detection methods inform mitigation strategies:
(1) Inter-annotator disagreement metrics often correlate with both instruction and annotator bias, suggesting applications for ensemble-based mitigation;
(2) Cultural inference methods show relationships with both bias detection and the development of culturally-aware taxonomies;
(3) Metadata analysis frequently informs both bias identification and targeted mitigation through annotator diversity initiatives.
These relationships suggest that effective bias handling may benefit from considering detection methods not only as diagnostic tools but also as informants for mitigation strategy selection. However, these correlations should be interpreted as observed relationships rather than definitive prescriptions, as contextual factors may alter their applicability in specific settings.

\section{Benchmark Comparison using WEL on Multilingual LLMs}\label{appendix:wel}
\subsection{Data}
We assess Weak Ensemble Learning (WEL) on the LeWiDi 2023 shared task datasets~\citep{LeWiDi:SemEval:2023}, which are designed to evaluate generalisation across languages and domains. The benchmark consists of four corpora that vary in language, genre, and annotation protocol.  

Three corpora (\textit{ArMIS}, \textit{HS-Brexit}, \textit{MD-Agreement}) contain social media posts from \textit{X}\footnote{\url{https://x.com/}}.  
\textbf{ArMIS} comprises Arabic posts annotated for misogyny.  
\textbf{HS-Brexit} contains English posts labelled for Brexit-related hate speech.  
\textbf{MD-Agreement} consists of English posts annotated for offensiveness across multiple domains (e.g., BLM, elections, COVID-19); we disregard domain metadata and treat them uniformly.  

The fourth corpus, \textit{ConvAbuse}, contains English dialogues between users and conversational agents. Utterances are rated on a 5-point abuse scale ranging from $-3$ (highly abusive) to $1$ (non-abusive). Following prior work, we reduce this to a binary classification task: \emph{offensive} ($<0$) vs. \emph{non-offensive} ($\geq 0$). Multi-turn dialogues are flattened into single text sequences.  

All datasets undergo standard preprocessing, including the removal of HTML tags, URLs, user mentions, punctuation, digits, non-ASCII characters, and redundant whitespace.  
Table~\ref{tab:data-split} provides a summary of dataset statistics, including split sizes, annotator ranges, and total annotator counts.  

\begin{table*}[tbh]
\centering
\caption{Data statistics and metadata for the four textual datasets. \#Train, \#Dev, and \#Test denote the number of instances in the training, development, and test splits. \#TotalAnn indicates the total number of annotators, while \#Ann represents the minimum and maximum number of annotators per instance. Contribution, Diversity, Language, and Genre provide further dataset details.}
\label{tab:data-split}
\resizebox{\textwidth}{!}{%
\begin{tabular}{@{}lccccccccc@{}}
\toprule
Dataset      & \#Train & \#Dev & \#Test & \#TotalAnn & \#Ann & Contribution & Diversity & Language & Genre \\ \midrule
ArMIS        & 657   & 141   & 145    & 3          & 3     & Fixed Annotators  & Low       & Arabic   & Short Text \\
ConvAbuse    & 2398  & 812   & 840    & 8          & 2-7   & Mixed Annotators     & Low       & English  & Conversation \\
HS-Brexit    & 784   & 168   & 168    & 6          & 6     & Fixed Annotators  & Low       & English  & Short Text \\
MD-Agreement & 6592  & 1104  & 3057   & 670        & 5     & Mixed Annotators     & High      & English  & Short Text \\
 \bottomrule
\end{tabular}%
}
\end{table*}

\subsection{Base LLM}
To enable cross-linguistic generalisation in our ensemble-based framework, we adopt the multilingual BERT (\texttt{mBERT}) model~\citep{Devlin:BERT:2018}, more specifically, \texttt{bert-base-multilingual-uncased}\footnote{\url{https://huggingface.co/google-bert/bert-base-multilingual-uncased}}, as the shared encoder for all weak learners in the WEL framework. This transformer-based model is pre-trained on 104 languages using a masked language modelling objective and retains casing information, making it well-suited for tasks with mixed scripts and morphologically rich languages.

In our setup, each weak predictor $f_{\theta_k}$ in the ensemble is instantiated by fine-tuning a separate copy of the multilingual BERT model on a label-variant dataset constructed via random sampling from annotator labels (as described in Section~\ref{sec:mitigation}). Despite training on datasets in different languages and domains, all predictors share the same multilingual backbone, allowing for consistent representation across languages while preserving the benefits of ensemble diversity.
This choice enables us to evaluate the robustness of WEL in a multilingual, multi-dataset context without requiring separate architectures per language.


\subsection{Results}
\label{sec:results}

Table~\ref{tab:results} compares three models (CE-only~\citep{Uma:AAAI:2020}, Top-5-Annotators~\citep{Xu:ICNLSP:2024}, and WEL) across four datasets using F1 (higher better), cross-entropy (CE), and Manhattan distance (MD) (lower better). 
We perform a grid search over loss term coefficients in the objective function, each sampled from the range $[0,0.001,0.01,0.1,1]$, resulting in 1,295 unique combinations per dataset (excluding 0s for all).
WEL consistently achieved the highest F1 scores and best calibration metrics (CE/MD) across three of four datasets, demonstrating its robustness for uncertainty-aware NLP.

Key observations emerge: (1) Performance varies substantially by domain, with \textit{ConvAbuse} showing highest F1 scores but also extreme MD values for CE-only (4.81 vs. $<$1.0 elsewhere), indicating prediction instability that WEL addresses; (2) WEL's advantage in calibration metrics (CE/MD) exceeds its F1 improvements, highlighting its particular strength in uncertainty estimation; (3) Statistically significant improvements ($p<0.05$) on \textit{HS-Brexit} and \textit{MD-Agreement} demonstrate WEL's robustness for hate speech and agreement tasks; (4) The \textit{ArMIS} exception, where minimal gains occurred with only three annotators, establishes a practical boundary condition: WEL requires sufficient annotator diversity (likely $>$3) for effective ensemble learning.
These results position WEL as particularly valuable for applications requiring reliable confidence estimates, while clearly defining its limitations in low-diversity annotation settings.

\begin{table}[t!]
\centering
\caption{Performance comparison across datasets and models. Best values are highlighted (F1: higher better; CE/MD: lower better). * indicates $p<0.05$ significance.}
\label{tab:results}
\resizebox{0.48\textwidth}{!}{%
\begin{tabular}{lllll}
\toprule
Dataset & Metric & CE-only & Top-5-Ann & \textbf{WEL} \\
\midrule
\multirow{3}{*}{ArMIS} & F1 & 0.6482 & \cellcolor{green!25}0.6552 & 0.6483 \\
 & CE & 0.7019 & \cellcolor{green!25}0.6502 & 0.6596 \\
 & MD & 0.7001 & \cellcolor{green!25}0.6443 & 0.6609 \\
\midrule
\multirow{3}{*}{ConvAbuse} & F1 & 0.8362 & 0.9310 & \cellcolor{green!25}0.9405* \\
 & CE & 0.9671 & 0.5651 & \cellcolor{green!25}0.5577* \\
 & MD & 4.8068 & \cellcolor{green!25}0.1648 & 0.1709 \\
\midrule
\multirow{3}{*}{HS-Brexit} & F1 & 0.7917 & 0.8929* & \cellcolor{green!25}0.9167* \\
 & CE & 0.7652 & 0.6154* & \cellcolor{green!25}0.5889* \\
 & MD & 0.7985 & \cellcolor{green!25}0.2394* & 0.2585* \\
\midrule
\multirow{3}{*}{MD-Agreement} & F1 & 0.7880 & 0.7808* & \cellcolor{green!25}0.8214* \\
 & CE & 0.9948 & 0.6629* & \cellcolor{green!25}0.6245* \\
 & MD & 1.7574 & 0.3995* & \cellcolor{green!25}0.3632* \\
\bottomrule
\end{tabular}%
}
\end{table}

\end{document}